# Analytical Characterization and Design Space Exploration for Optimization of CNNs

Rui Li
lirui@cs.utah.edu
University of Utah
Salt Lake City, Utah, USA

Yufan Xu
yf.xu@utah.edu
University of Utah
Salt Lake City, Utah, USA

Aravind Sukumaran-Rajam
a.sukumaranrajam@wsu.edu
Washington State University
Pullman, Washington, USA

Atanas Rountev
rountev@cse.ohio-state.edu
Ohio State University
Columbus, Ohio, USA

P. Sadayappan
saday@cs.utah.edu
University of Utah
Salt Lake City, Utah, USA

## ABSTRACT

Moving data through the memory hierarchy is a fundamental bottleneck that can limit the performance of core algorithms of machine learning, such as convolutional neural networks (CNNs). Loop-level optimization, including loop tiling and loop permutation, are fundamental transformations to reduce data movement. However, the search space for finding the best loop-level optimization configuration is explosively large. This paper develops an analytical modeling approach for finding the best loop-level optimization configuration for CNNs on multi-core CPUs. Experimental evaluation shows that this approach achieves comparable or better performance than state-of-the-art libraries and auto-tuning based optimizers for CNNs.

**ACM Reference Format:**
Rui Li, Yufan Xu, Aravind Sukumaran-Rajam, Atanas Rountev, and P. Sadayappan. 2021. Analytical Characterization and Design Space Exploration for Optimization of CNNs. In *Proceedings of the 26th ACM International Conference on Architectural Support for Programming Languages and Operating Systems (ASPLOS '21), April 19–23, 2021, Virtual, USA.* ACM, New York, NY, USA, 15 pages. https://doi.org/10.1145/3445814.3446759

## 1 INTRODUCTION

Convolutional Neural Networks (CNNs) have had transformative impact on several domains including image/video classification,

language processing, genetic analysis, etc. CNNs are computationally very demanding. Therefore there has been tremendous interest in optimized implementation of the CNN stages needed in Deep Neural Network (DNN) pipelines. CNN stages of varied shapes and sizes are needed even within a single DNN pipeline.

Since the cost of data movement dominates the cost of floating-point arithmetic computations on all current hardware platforms, loop tiling is a crucial transformation for the development of optimized code for CNN. However, a fundamental challenge is the explosive size of the space of possible tiled loop variants for the CNN computation:

$$Out[n, k, h, w] = \sum_{c,r,s} In[n, c, h + r, w + s] * Ker[k, c, r, s] \quad (1)$$

The computation can be expressed as a 7-dimensional loop nest, with one loop per index. Allowing for any order of accumulation of additive contributions for each result tensor element, all 7 loops are fully permutable and hence fully tileable with hyper-rectangular tiles. Considering a three-level memory hierarchy, up to three levels of tiling may be appropriate, leading to an explosively large search space with three groups of 7 tiling loops, with 7! possible permutations of the tiling loops within each group, i.e., $1.28 \times 10^{11}$ configurations. Further, for each configuration of tiling loops, a very large number of possible choices exist for the tile sizes, resulting in an explosive number of alternatives from which to select.

To the best of our knowledge, all previously developed approaches for CNN optimization have used heuristics and/or empirical auto-tuning to search *a limited subset of the explosive space* of permutations and tile size choices [6, 20, 34]. This is a fundamental limitation to achieving consistently high performance across the wide range of CNN instances used in DNN pipelines. We aim to solve this problem in a principled and comprehensive way. To achieve this, we develop the first approach that models analytically the data movement for any CNN stage in a multi-level memory hierarchy. Using this model, we show how to explore *the entire search space*, looking for the configuration that minimizes the bandwidth-scaled data movement in the limiting level of the memory hierarchy. The insight of our approach, which differentiates it from previous CNN optimization efforts, is that analytical modeling and reasoning enable dramatic pruning of the space of permutations and tile sizes, reducing it to a small number of non-linear optimization problems that can be solved by off-the-shelf solvers. This paper targets multicore CPUs, but the analytical machinery

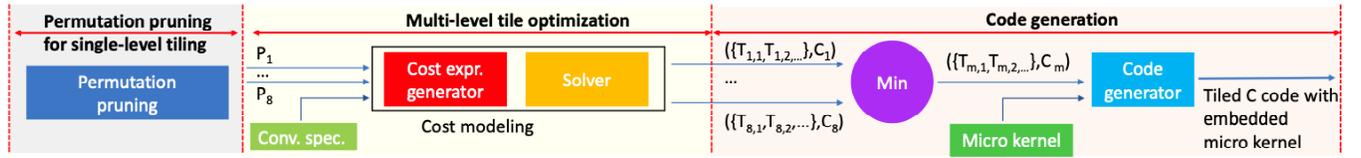

**Figure 1: MOpt Overview**

is applicable to targets such as GPUs, TPUs, FPGAs, and spatial arrays of accelerators.

Our modeling approach addresses a key limitation of existing efforts for CNN optimization. To demonstrate its utility, in this paper we combine this modeling with our custom code generator to achieve CNN performance that matches or exceeds the performance possible with state-of-the-art approaches. In the long run, our techniques provide a critical building block for these existing approaches, allowing them to overcome one of their fundamental limitations. This existing work falls in the following three categories.

**Libraries of optimized functions:** Tuned vendor libraries are currently the primary means of achieving high performance for most applications using CNNs. Applications are typically developed by composing operators in a high-productivity framework such as PyTorch or TensorFlow, with the frameworks mapping the execution of the operators to invocation of tuned library function calls. Although vendor libraries can achieve very good performance, we demonstrate through our experimental evaluation of Intel's state-of-the-art oneDNN library that there is scope for improvement if wider exploration of the search space is be undertaken using the approach proposed in this paper (the discussion in Sec. 12 elaborates on this).

**Auto-tuning and ML-based tuning:** One of the most successful recent efforts in optimizing tensor computations has been TVM [6]. TVM uses a combination of auto-tuning (actual execution of candidate code variants on the target platform) and a dynamically trained Machine Learning model to guide the design-space exploration. However the enormous search space poses a problem and manual expertise is required to set up optimization scripts that control the search space. We present experiments demonstrating the greater effectiveness of our new approach over TVM's auto-tuning over a constrained search space. By combining the model-driven comprehensive design space exploration from our work with the auto-tuning framework in TVM, further improvement in performance is feasible (the discussion in Sec. 12 elaborates on this).

**Polyhedral compilers:** Such compilers incorporate powerful transformations for affine programs [4, 5, 8, 36]. The CNN computation in Eq. 1 is affine and can be automatically tiled and optimized by this approach. However, the performance achieved by state-of-the-art polyhedral compilers is very far from that provided by vendor libraries or by auto-tuning-based code generators such as TVM [6]. These compilers face a fundamental challenge: they must separate the key consideration of tile size optimization—inherently non-linear—from the choice of loop transformations. The only recourse is to use an outer auto-tuning loop that explores a limited space of tile sizes, and an inner loop that generates code for them [2, 5, 11, 18, 27, 35, 36]. Our approach can be generalized for analytical modeling of data movement in a class of tiled tensor computations and incorporated into polyhedral compilers, thereby overcoming this fundamental limitation. (Sec. 12 elaborates on this).

**Contributions:** The paper makes the following contributions:

● It develops, to the best of our knowledge, the first comprehensive analytical modeling for data movement volume for multi-level tiled CNN execution on a system with a multi-level memory hierarchy, covering the full space of permutations and tile sizes. While the modeling approach is used in the context of multicore CPUs, it can also be used for CNN optimization on other platforms, such as GPUs, FPGAs, distributed-memory systems, and accelerator arrays.

● It presents the first analysis that exploits algebraic properties of the analytical expressions for data-movement volume to dramatically prune the number of distinct cases *from thousands to only eight* in order to find the global optimum in the entire space of tile-loop permutations for a single-level tiled CNN. The factor of reduction in the search space that is enabled by this algebraic analysis is exponentially higher for multi-level tile-size optimization.

● It demonstrates the use of the new analytical modeling and optimization approach through the generation of high-performance multicore CPU code for three CNN benchmarks, including all CNN stages of MobileNet [14], ResNet-18 [13], and Yolo9000 [29]. The achieved performance is comparable to or better than both the state-of-the-art CNN library (Intel's oneDNN [25]) and the state-of-the-art framework for auto-tuned code generation (TVM [6]).

## 2 OVERVIEW

### 2.1 System Overview

Fig. 1 shows the components of the *MOpt* system (*M*odeling-based *Opt*imizer) for generating optimized CNN code for multicore processors, based on a novel comprehensive design-space exploration approach for tile-loop optimization. The leftmost component represents a conceptual methodology for pruning the space of possible permutations of tile-loops for single-level tiling. This methodology uses analytical modeling of data movement volume to identify a very small subset—containing only 8 elements—of the full space of tile-loop permutations, guaranteed to contain an optimal configuration that minimizes data volume for tiled execution. The rest of this section highlights the key ideas behind this modeling, while Sec. 3 and 4 provide a more detailed description.

The right portion of the figure shows the tool components for code generation for a specific CNN. From the insights provided by the modeling methodology, together with the specific sizes of the kernel and input/output of the CNN, a set of constrained non-linear optimization problems are automatically generated. These problems capture the search for optimal tile sizes for multi-level tiling (Sec. 5). The optimization problems are then solved using an

```
// Ni/Nj/Nk  are  perfect  multiples  of  Ti/Tj/Tk
for( it = 0;  it < Ni;  it+=Ti)
 for( jt = 0;  jt < Nj;  jt+=Tj)
  for( kt = 0;  kt < Nk;  kt+=Tk)
   for( i = 0;  i < Ti;  i++)
    for( j = 0;  j < Tj;  j++)
     for( k = 0;  k < Tk;  k++)
      C[ i*it ][ j+jt ]*=
      A[ i*it ][ k+kt ]*B[k+kt][ j+jt ];
```

**Listing 1: Single-level tiled matrix multiplication**

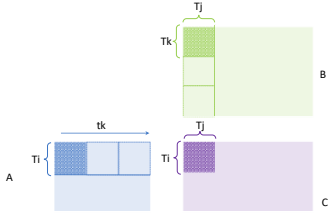

**Figure 2: Data reuse in tiled matrix multiplication**

off-the-shelf non-linear solver (we use AMPL [9] with Ipopt [37]) to produce optimal tile sizes $T_{i,j}$ and data movement costs $C_i$ (here $j$ ranges over the levels of the memory hierarchy). The best solution gives the tile sizes and tile-loop permutation to be used to generate customized C code for the CNN stage, with tile loops surrounding a CNN microkernel that implements register-tiling using vector intrinsics.

## 2.2  Key Ideas for Analytical Modeling

We use the simpler example of matrix multiplication to explain the main ideas behind the new approach to comprehensive design space exploration for tiled CNN optimization. For the CNN computation, the analytical cost functions are more general than for matrix multiplication, but have a similar structure. Furthermore, the reasoning to derive these functions and to optimize tile sizes based on them is also similar. Listing 1 shows one possible version of single-level tiled code for matrix-multiplication (there are $6 \times 6 = 36$ possible permuted variants, with 6 possible permutations for the intra-tile loops and 6 possible permutations of the tiling loops).

Consider the data footprint of a single tile from Listing 1. This footprint is the sum of the volumes of the data slices accessed by the three arrays $A$, $B$, and $C$, respectively $T_iT_k$, $T_jT_k$, and $T_iT_j$. This is illustrated in Fig. 2. Among all possible combinations of tile sizes chosen such that the total data-footprint does not exceed cache capacity, we want to find the one(s) achieving minimal data movement between main memory and cache:

$$T_iT_k + T_jT_k + T_iT_j \leq C \qquad (2)$$

As is the case with much of the prior work on analytical modeling of cache misses for loop computations [12][3][16], we only model cold misses (first access of data) and capacity misses but not conflict misses arising from finite set-associativity of caches. We demonstrate through experimental evaluation that this idealized model of cache behavior is very effective in tile optimization for CNNs.

Consider the iterations of the innermost tiling loop $kt$. As $kt$ is changed, and different tiles are executed, we can observe (Fig. 2) that the accessed data slices are completely distinct (i.e., without any reuse of data between tiles) for $A$ and $B$, whereas exactly the same data slice of $C$ is used for all the tiles. The total volume of data movement between main memory and cache for the complete execution of the innermost tiling loop $kt$ is $DV_{kt}^A = T_iN_k$ and $DV_{kt}^B = T_jN_k$ for arrays $A$ and $B$, respectively. For $C$, since the same data slice $C[it:it+T_i-1][jt:jt+T_j-1]$ is repeatedly accessed for each value of the tile-loop iterator $kt$, with a fully associative cache each data element will only be brought in once from memory.

The combined data volume for all three arrays, $DV_{kt}$, is as follows (the factor of 2 associated with the data volume for $C$ is due to the need to move each element in both directions, first from memory to cache and finally back from cache to memory):

$$DV_{kt} = DV_{kt}^A + DV_{kt}^B + DV_{kt}^C = T_iN_k + T_jN_k + 2T_iT_j$$

The modeling of total data movement volume between memory and cache for the execution of the innermost $kt$ tile-loop was facilitated by the fact that two of the arrays did not have any inter-tile data reuse, while the third one had complete inter-tile data reuse of a slice of data that was small enough to fit in the cache. As we attempt to analyze the volume of data movement through the outer two tiling loops, the data footprints of the arrays increase and the analysis of hits and misses becomes very complicated, with many combinations of possibilities depending on the chosen tile sizes.

A key to developing our analytical parametric modeling approach is the recognition that for the purpose of tile-size optimization, we do not need to accurately model data-movement volume for all possible tile sizes, but it is sufficient to carry out such modeling for those tile sizes that effectively utilize the available capacity of the cache/scratchpad. We therefore assume that the collective data footprint of two adjacent tiles will exceed the cache capacity – if not, the chosen tile sizes are too small and wasteful and should be increased to make better use of the available capacity. Under such an assumption, we can continue the parametric analysis of data volume for the entire execution of the tiled matrix multiplication algorithm. For any tiling loop, we have two possibilities with respect to any array: the loop iterator is either used in the indexing of the array (it is a *present* index), or it is not used and thus is an *absent* index (e.g., tile-loop iterator $it$ does not affect the accessed elements of array $B[k][j]$ because $i$ is an *absent* index for $B$). If the tile-loop iterator is a *present* index, the data slice accessed for each value of the iterator is distinct, and the total accessed data volume over the execution of the tile loop is the product of the number of tile-loop iterations and the data volume corresponding to the inner nested loops. Even if the tile-loop iterator is an *absent* index, if the data footprint of the slice accessed by inner loops has exceeded cache capacity, the total data movement is again the product of the number of tile-loop iterations and the data volume accessed in execution of the inner loops. Based on these observations, the following cost expression applies to the two innermost tile-loops:

$$DV_{jt,kt} = \frac{N_j}{T_j}DV_{kt} = \frac{N_j}{T_j}T_iN_k + N_jN_k + 2T_iN_j$$

Similarly,

$$\begin{aligned} DV_{it,jt,kt} &= \frac{N_i}{T_i}DV_{jt,kt} = \frac{N_i}{T_i}\left(\frac{N_j}{T_j}T_iN_k + N_jN_k + 2T_iN_j\right) \\ &= N_iN_jN_k\left(\frac{1}{T_j} + \frac{1}{T_i} + \frac{2}{N_k}\right) \end{aligned} \qquad (3)$$

Given specific values for $N_i$, $N_j$, $N_k$, the parametric expression in Eq. 3 can be minimized subject to the capacity constraints in Eq. 2. However, this is only one of 6 permutations of the tiling loops, and we desire the combination of tile-loop permutation and tile sizes that minimize total data movement between memory and cache.

```
for(n = 0; n < Nn; n++)
 for(k = 0; k < Nk; k++)
  for(c = 0; c < Nc; c++)
   for(r = 0; r < Nr; r++)
    for(s = 0; s < Ns; s++)
     for(h = 0; h < Nh; h++)
      for(w = 0; w < Nw; w++)
       Out[n][k][h][w] +=
        In[n][c][h+r][w+s]*Ker[k][c][r][s]
```

**Listing 2: CNN loops**

```
for(nt = 0; nt < Nb; nt+=Tn)
 for(kt = 0; kt < Nk; kt+=Tk)
  for(ct = 0; ct < Nc; ct+=Tc)
   for(rt = 0; rt < Nr; rt+=Tr)
    for(st = 0; st < Ns; st+=Ts)
     for(ht = 0; ht < Nh; ht+=Th)
      for(wt = 0; wt < Nw; wt+=Tw)
       CNNTile(nt,kt,ct,rt,st,ht,wt);
```

**Listing 3: CNN with single-level tiling**

When this modeling is generalized to the CNN computation (as described in the next section), a brute-force enumeration and solution of a constrained optimization problem for each possible tile-loop permutation leads to a huge number of cases. For example, for multi-level tiling of the 7-dimensional loop nest for CNN, with 4 levels of tiling loops (register-tiling, L1, L2, and L3 cache), the number of cases is $(7!)^4$, i.e., over 645 trillion cases. However, as we elaborate in Sec. 4, algebraic reasoning can be used to reduce the total number of parametric symbolic expressions to be considered for modeling all tile-loop permutations at one level of tiling for CNN from 7! (i.e., 5040) to only 8. This massive pruning is possible because of algebraic reasoning about equivalence or dominance (guaranteed to find a better or equally good solution) of all remaining 5032 cases by these 8 constrained optimization problems.

## 3 ANALYTICAL MODELING FOR SINGLE-LEVEL TILING

Given a specific permutation of the tile-loops for a single level of tiling of the CNN computation, we aim to develop a parametric expression for the total volume of data movement (as a function of tile sizes) between main memory and an idealized fully-associative LRU cache with a capacity of $C$ words and unit line-size. In the next section, we present a pruning strategy to dramatically reduce the number of tile-loop permutations to be considered in solving the tile-optimization problem. Given the original CNN code in Listing 2, Listing 3 shows one particular single-level tiled version.[1] We will use $\vec{p} = \langle i_7, \ldots, i_1 \rangle$ to denote a particular permutation of the tile-loop iterators $nt, kt, \ldots$ in the tiled code, where $i_1$ is the innermost tile-loop iterator in the tile-loop nest. The corresponding tile sizes for a particular tiled version will be denoted by $\vec{T} = \langle T_7, \ldots, T_1 \rangle \in \mathbb{N}^7$. Here each tile size $T_j$ is such that $1 \leq T_j \leq N_j$ where $N_j$ is the corresponding problem size. We assume that each problem size $N_j$ is a multiple of the corresponding tile size $T_j$. This assumption is used only for the presentation of cost modeling; the actual code generation handles the general case of partial tiles. A *tiling configuration* is a pair $\langle \vec{p}, \vec{T} \rangle$.

In the execution, the iterators from $\vec{p}$ will be instantiated with concrete values. Each such instance is an iteration vector and will be denoted by $\vec{i} \in \mathbb{N}^7$. In any such $\vec{i}$, the value of iterator $i_j$ is always a multiple of the corresponding tile size $T_j$. To simplify the discussion, in our cost modeling we will normalize $i_j$ in $\vec{i}$ by $T_j$. Thus, the $j$-th element of $\vec{i}$ now takes values in the set $\{0, 1, \ldots, N_j/T_j\}$. Execution of the code defined by a configuration $\langle \vec{p}, \vec{T} \rangle$ corresponds to a sequence of tiles defined by a lexicographic order of all vectors $\vec{i}$. A key component of our modeling is an analytical description of the amount of data movement in executing two consecutive tiles.

### 3.1 Overview of Modeling of Inter-Tile Data Reuse and Total Data Movement

Given $\vec{p} = \langle i_7, \ldots, i_1 \rangle$, we construct an analytical expression to model the amount of data movement when the corresponding tiled execution occurs. Note that the expression is parametric in the tile sizes $\vec{T}$ and will later be used to define a constrained optimization problem in which the objective function is this cost expression and the unknowns are the tile sizes in $\vec{T}$. Thus, for any code version (as defined by a loop permutation $\vec{p}$), the solution of this optimization problem provides concrete tile sizes to minimize the cost expression.

The modeling analysis is done separately for each of the three arrays $In$, $Out$, and $Ker$. For any array $A$, let $R_A$ be innermost (i.e., rightmost) position in $\vec{p}$ of an iterator that occurs in the array reference for $A$. For example, suppose $\vec{p} = \langle \ldots, ct, nt \rangle$. For array reference $Out[n, k, h, w]$ from the original code we have $R_{Out} = 1$, since in the tiled code this reference becomes $Out[n + nt, k + kt, h + ht, w + wt]$ which contains $nt$, and $nt$ is in position 1 in $\vec{p}$. For array reference $In[n, c, h + r, w + s]$, both $nt$ and $ct$ occur in the tiled code, but $nt$ occurs at position 1 in $\vec{p}$ (i.e., in the innermost/rightmost position) and thus $R_{In} = 1$. Finally, for $Ker[k, c, r, s]$ we have $R_{Ker} = 2$ since $ct$ occurs at position 2 in $\vec{p}$.

Consider a tile with tile sizes $T_n$, $T_k$, $T_c$, $T_r$, $T_s$, $T_h$, $T_w$. The execution of the tile will access a 4-D slice of $T_n T_k T_h T_w$ elements of $Out[n, k, h, w]$ and $T_k T_c T_r T_s$ elements of $Ker[k, c, r, s]$. For $In[n, c, h + r, w + s]$, the data slice accessed in the tile will have $T_n T_c (T_h + T_r - 1)(T_w + T_s - 1)$ elements. This is because the index expression $w + s$ takes $T_w + T_s - 1$ distinct values in a contiguous range as $w$ varies over some contiguous range of $T_w$ values and $s$ ranges over a range of $T_s$ values. The capacity constraint specifying that the total data footprint must not exceed cache capacity is:

$$T_n T_c (T_h + T_r - 1)(T_w + T_s - 1) + T_k T_c T_r T_s + T_n T_k T_h T_w \leq C \quad (4)$$

As illustrated in Sec. 2 with the matrix-multiplication example, the analytical modeling of data volume for execution of the CNN loop nest for a specific tile-loop permutation is done by an inner to outer traversal of the tile-loops. Starting with the inner-most tile loop, that loop's index is either absent or present in the tensor's index expressions. For example, consider the particular tile-loop order shown in Listing 3. The innermost tile-loop corresponds to loop index $wt$, which is an *absent* iterator for $Ker$ and a *present* iterator for $In$ and $Out$. This means that for $Ker$ the data slices accessed for successive tiles as we step through the $wt$ tile-loop will be exactly the same, i.e., full inter-tile data reuse is achieved.

---

[1]To simplify the presentation, we do not show stride/dilation, but the methodology is applicable to the general case.

In contrast, completely distinct data slices of $Out$ are accessed by the different tiles that are executed as $wt$ is varied, i.e., there is absolutely no data reuse across the tiles. For $In$, the original indexing expression involving $w$ is of the form $w + s$. Hence there is some partial overlap of the data slices accessed by successive tiles as $wt$ iterates (as detailed below).

For any permutation $\vec{p}$, for the innermost tile-loop there is complete data reuse between successive tiles if that iterator is *absent* in a tensor's index expressions, and no reuse or partial reuse for any tensor where the index is *present*. Further, after the execution of all tiles in the innermost tile-loop, eviction of data from previous tiles should occur for any tensor with that index *present*. This is a consequence of our choice in only modeling data-movement volume for tile sizes that are sufficiently large so that cache capacity is not wasted (i.e, the combined tile footprint of two adjacent tiles always exceeds cache capacity). Thus, for any tensors with the innermost tile loop index being *present*, no data reuse is possible at any outer tiling loops even if that outer index is *absent*.

## 3.2 Cost Expressions for Data Movement

Based on these properties, there are two cases for the cost computation. The first case is for arrays $Out$ and $Ker$, as well as for $In$ when the iterator at position $R_{In}$ is $nt$ or $ct$. Here the cost computation simply considers the number of pairs of consecutive iteration vectors $\vec{i}$ and $\vec{i'}$ in the lexicographic order such that the value at position $R_A$ changes from the first to the second vector. In all such cases, the second tile accesses a completely different slice of the corresponding array $A$. Thus, the amount of data movement is the number $\prod_{R_A \le j \le 7} \frac{N_j}{T_j}$ of such pairs multiplied by the tile footprint for that array.

As discussed earlier, for $Out$ the tile footprint is $T_n T_k T_h T_w$ and for $Ker$ this footprint is $T_k T_c T_r T_s$. For array $In$, the footprint is $T_n T_c (T_h + T_r - 1)(T_w + T_s - 1)$. Multiplying this footprint with the number of pairs of consecutive tiles for which data movement occurs (as defined above) gives the complete data volume for a particular loop permutation $\vec{p}$.

The second case is for $In[n, c, h + r, w + s]$ when the iterator at position $R_{In}$ is $wt$, $ht$, $st$, or $rt$. Consider one execution of the loop for this iterator. Each time the iterator changes, there is partial reuse across consecutive tiles. As a result, the inter-tile movement cost along the corresponding data dimension is the tile size for the iterator. For example, if the iterator at position $R_{In}$ is $wt$, the tile footprint in that data dimension is $T_s + T_w - 1$, but due to partial overlap between tiles the actual amount of new data in that data dimension is $T_w$. For one execution of the $wt$ loop, there are $N_w/T_w - 1$ such iterator changes. Thus, the cost is $T_w(N_w/T_w - 1) = N_w - T_w$. The number of times this cost is incurred is determined by the loops surrounding $wt$, and is the product of $N_j/T_j$ for the positions $j$ around $R_{In}$.

More generally, we have a cost term which is the product of $\prod_{R_{In} < j \le 7} \frac{N_j}{T_j}$ and one of the following:

- $T_n T_c (T_h + T_r - 1)(N_w - T_w)$ when $wt$ is at $R_{In}$
- $T_n T_c (T_h + T_r - 1)(N_s - T_s)$ when $st$ is at $R_{In}$
- $T_n T_c (N_h - T_h)(T_w + T_s - 1)$ when $ht$ is at $R_{In}$
- $T_n T_c (N_r - T_r)(T_w + T_s - 1)$ when $rt$ is at $R_{In}$

We also have a second term which captures data movement cost when the very first iteration of that loop occurs. For this iteration there is no reuse from the previous tile, and the cost of the entire tile footprint is incurred. This cost is the product of $\prod_{R_{In} < j \le 7} \frac{N_j}{T_j}$ and $T_n T_c (T_h + T_r - 1)(T_w + T_s - 1)$.

# 4 PRUNING CONFIGURATIONS: SINGLE-LEVEL TILING

Sec. 3 presented symbolic expressions for total data volume as a function of parametric tile sizes, for any given permutation of the tile-loops. There are 7! possible permutations for the seven tile loops for a single level of cache, and $(7!)^L$ permutations for $L$ levels of cache. In this section, we show that massive pruning of the search space is possible via algebraic analysis that reduces the number of permutations to be considered to just 8 of the 7! = 5040 total permutations of the seven tile-loops. This is done by proving that the solution to one of these eight optimization problems is guaranteed to be as good as or better than any solutions for the remaining 5032 cases.

The identification of the pruned subset of tile-loop permutations is done via an inner-to-outer analysis of tiling loops and reasoning about the implications on total data movement cost, for different choices for tile-loop indices made at each level. The array indexing structure for the CNN computation is such that each of the seven loop indices is *present* in exactly two of the three tensors and *absent* in one tensor: $w$, $h$, and $n$ are all present for $In$ and $Out$, but absent for $Ker$; $s$, $r$, and $c$ are present for $In$ and $Ker$, but absent for $Out$; $k$ is present for $Ker$ and $Out$ but absent for $In$. As per the analysis in the previous section, the total data movement cost for two of the three arrays will be fully determined just from the choice of the innermost tile-loop. The rest of this section describes these cases and summarizes the final result of this reasoning.

**Innermost** $wt$: If we choose the innermost tile-loop to be $wt$, the data movement volume for the seven tiling loops will be $\frac{N_n}{T_n} \frac{N_k}{T_k} \frac{N_c}{T_c} \frac{N_r}{T_r} \frac{N_s}{T_s} \frac{N_h}{T_h} T_n T_c (T_h + T_r - 1)(N_w + T_s - 1)$ for $In$ and $2 \frac{N_n}{T_n} \frac{N_k}{T_k} \frac{N_c}{T_c} \frac{N_r}{T_r} \frac{N_s}{T_s} \frac{N_h}{T_h} \frac{N_w}{T_w} T_n T_k T_h T_w$ for $Out$ (the factor of 2 is due to the need to read and write each element of $Out$).

The order of the six surrounding tile-loops will not affect the total data movement cost of $In$ and $Out$, but will affect the data movement cost for $Ker$. As per the analysis in Sec. 3, the expression for data movement for $Ker$ is a product of the tile footprint's volume $(T_k T_c T_r T_s)$ and the product of $N_j/T_j$ for all tiles from the first *present* iterator and all surrounding iterators. The volume will be minimized if all *absent* indices are lower in the nesting order than all *present* indices. This is achieved by placing the tile-loops for *absent* indices $ht$ and $nt$ (in either order) in a band just above $wt$, with the tile-loops for *present* indices $kt$, $ct$, $rt$, and $st$ in a band (in any order) above the tile-loops for $ht$ and $nt$. We will use the notation $\langle\{kt, ct, rt, st\}, \{nt, ht\}, wt\rangle$ to denote the set of tile-loop configurations described above: innermost tile-loop for $wt$, surrounded by a band of two tile-loops for $nt$ and $ht$ (in either order), and an outermost band of tile-loops for indices $kt$, $ct$, $rt$, $st$, in any relative order among those four tile-loops. Note that this notation represents a set of $4! \times 2! = 48$ iterator permutations; however, all elements of this set are equivalent with respect to the cost model, as their cost expressions are exactly the same. When exploring the search space, one arbitrary representative of this set will be chosen and will be subjected to non-linear optimization. The same applies for the other seven cases described below: each case defines a set of cost-equivalent permutations, and one arbitrary representative of the set is selected for tile size optimization.

The parametric expression for the total data movement cost for any configuration in set $\langle \{kt, ct, rt, st\}, \{nt, ht\}, wt \rangle$, e.g., $\langle kt, ct, rt, st, nt, ht, wt \rangle$ is:

$$DV_{kt,ct,rt,st,nt,ht,wt} = \frac{N_k}{T_k} \frac{N_c}{T_c} \frac{N_r}{T_r} \frac{N_s}{T_s} [T_k T_c T_r T_s +$$
$$\frac{N_n}{T_n} \frac{N_h}{T_h} (2 \frac{N_w}{T_w} T_n T_k T_h T_w + T_n T_c (T_h + T_r - 1)(N_w + T_s - 1))] \quad (5)$$

The solution of a constrained optimization problem to minimize the expression in Eq. 5, subject to the capacity constraint in Eq. 4 will find the lowest possible data volume among all possible permutations with $wt$ as the innermost tiling loop.

**Innermost $ht$:** The analysis for tile-loop configurations with $ht$ at the innermost position can be done similarly to the case with $wt$ being innermost. The minimal possible data movement will be achieved with any arbitrary member of the set $\langle \{kt, ct, rt, st\}, \{nt, wt\}, ht \rangle$, e.g., $\langle kt, ct, rt, st, nt, wt, ht \rangle$:

$$DV_{kt,ct,rt,st,nt,wt,ht} = \frac{N_k}{T_k} \frac{N_c}{T_c} \frac{N_r}{T_r} \frac{N_s}{T_s} [T_k T_c T_r T_s +$$
$$\frac{N_n}{T_n} \frac{N_w}{T_w} (2 \frac{N_h}{T_h} T_n T_k T_h T_w + T_n T_c (T_w + T_s - 1)(N_h + T_r - 1))]$$

**Innermost $st$:** Since $st$ is present for $In$ and $Ker$, the data movement costs for these two tensors will be independent of the permutations of the remaining outer tile-loop indices:

$$DV_{...,st}^{Ker} = \frac{N_n}{T_n} \frac{N_k}{T_k} \frac{N_c}{T_c} \frac{N_r}{T_r} \frac{N_s}{T_s} \frac{N_w}{T_w} \frac{N_h}{T_h} T_k T_c T_r T_s$$
$$DV_{...,st}^{In} = \frac{N_n}{T_n} \frac{N_k}{T_k} \frac{N_c}{T_c} \frac{N_r}{T_r} \frac{N_w}{T_w} \frac{N_h}{T_h} \times$$
$$T_k T_c (T_h + T_r - 1)(T_w + N_s - 1)$$

The data-movement cost for $Out$ will depend on the permutation of the outer tile-loops. The lowest cost is obtained when the *absent* indices for $Out$ are placed immediately above $st$. The absent indices for $Out$ are $ct$ and $rt$. Any permutation in the set $\langle \{nt, kt, ht, wt\}, \{ct, rt\}, st \rangle$ will achieve the lowest possible data movement cost for $Out$:

$$DV_{...,st}^{Out} = 2 \frac{N_n}{T_n} \frac{N_k}{T_k} \frac{N_h}{T_h} \frac{N_w}{T_w} T_n T_k T_h T_w$$

The optimization problem for any permutation in the set $\langle \{nt, kt, ht, wt\}, \{ct, rt\}, st \rangle$ is to minimize the sum of these three $DV$ cost expressions subject to the constraint in Eq. 4.

**Innermost $rt$:** The reasoning for this case is similar to the case for innermost $st$. The best permutations are in set $\langle \{nt, kt, ht, wt\}, \{ct, st\}, rt \rangle$. For them, the data movement cost is as follows:

$$DV_{...,rt}^{Out} = 2 \frac{N_n}{T_n} \frac{N_k}{T_k} \frac{N_h}{T_h} \frac{N_w}{T_w} T_n T_k T_h T_w$$
$$DV_{...,rt}^{Ker} = \frac{N_n}{T_n} \frac{N_k}{T_k} \frac{N_c}{T_c} \frac{N_r}{T_r} \frac{N_s}{T_s} \frac{N_w}{T_w} \frac{N_h}{T_h} T_k T_c T_r T_s$$
$$DV_{...,rt}^{In} = \frac{N_n}{T_n} \frac{N_k}{T_k} \frac{N_c}{T_c} \frac{N_s}{T_s} \frac{N_w}{T_w} \frac{N_h}{T_h} \times$$
$$T_k T_c (T_h + N_r - 1)(T_w + T_s - 1)$$
$$DV_{...,rt} = DV_{...,rt}^{Out} + DV_{...,rt}^{Ker} + DV_{...,rt}^{In}$$

**Innermost $kt$:** In this case the data movement volume will be $2 \frac{N_n}{T_n} \frac{N_k}{T_k} \frac{N_c}{T_c} \frac{N_r}{T_r} \frac{N_s}{T_s} \frac{N_w}{T_w} T_n T_k T_h T_w$ for $Out$ and $\frac{N_n}{T_n} \frac{N_k}{T_k} \frac{N_c}{T_c} \frac{N_r}{T_r} \frac{N_s}{T_s} \frac{N_w}{T_w} T_k T_c T_r T_s$ for $Ker$. Since $kt$ is absent in $In$, the next surrounding loop will contain an iterator that is present in $In$. This next iterator uniquely determines the cost function. The six cases for this choice can be separated in two groups: $\{wt, ht, st, rt\}$ and $\{nt, ct\}$. As discussed shortly, the second group of choices can be ignored. Any choice from the first group gives rise to a different cost expression; thus, each of those 4 cases has to be solved separately. Together with the 4 cases described earlier (i.e., innermost loop is $wt$, $ht$, $st$, or $rt$), this gives us the 8 overall cases mentioned previously.

The cost functions for the first group are similar to those discussed earlier. For example, the cost for $\langle \ldots, wt, kt \rangle$ is similar to the one for $\langle \ldots, wt \rangle$, but now a factor $\frac{N_k}{T_k}$ is missing because $kt$ is the innermost loop and does not affect $In$.

Now consider the second group $\{nt, ct\}$ of choices—for example, $\langle \ldots, nt, kt \rangle$. Compare this cost with the corresponding one for configuration $\langle \ldots, wt, kt \rangle$. It is easy to show that the only difference is a factor of $\frac{N_w}{T_w}(T_w + T_s - 1)$ in the cost for $\langle \ldots, nt, kt \rangle$, which is changed to $N_w + T_s - 1$ in the cost for $\langle \ldots, wt, kt \rangle$. Since $\frac{N_w}{T_w} \geq 1$, the cost for $\langle \ldots, nt, kt \rangle$ will never be lower than the one for $\langle \ldots, wt, kt \rangle$. Thus, $nt$ (and, similarly, $ct$) should not be chosen for the loop immediately surrounding the innermost loop $kt$.

For completeness, below are the details of the cost expressions for the four relevant cases. Based on different choices for the second innermost iterator, the data movement volume expression is as follows:

For permutation $\langle \{nt, ct, ht, rt, st\}, wt, kt \rangle$

$$DV_{...,wt,kt}^{Out} = 2 \frac{N_n}{T_n} \frac{N_k}{T_k} \frac{N_c}{T_c} \frac{N_r}{T_r} \frac{N_s}{T_s} \frac{N_h}{T_h} \frac{N_w}{T_w} T_n T_k T_h T_w$$
$$DV_{...,wt,kt}^{Ker} = \frac{N_n}{T_n} \frac{N_k}{T_k} \frac{N_c}{T_c} \frac{N_r}{T_r} \frac{N_s}{T_s} \frac{N_w}{T_w} \frac{N_h}{T_h} T_k T_c T_r T_s$$
$$DV_{...,wt,kt}^{In} = \frac{N_n}{T_n} \frac{N_c}{T_c} \frac{N_r}{T_r} \frac{N_s}{T_s} \frac{N_h}{T_h} \times$$
$$T_n T_c (T_h + T_r - 1)(N_w + T_s - 1)$$
$$DV_{...,wt,kt} = DV_{...,wt,kt}^{Out} + DV_{...,wt,kt}^{Ker} + DV_{...,wt,kt}^{In}$$

For permutation $\langle \{nt, ct, wt, rt, st\}, ht, kt \rangle$

$$DV_{...,ht,kt}^{Out} = 2 \frac{N_n}{T_n} \frac{N_k}{T_k} \frac{N_c}{T_c} \frac{N_r}{T_r} \frac{N_s}{T_s} \frac{N_h}{T_h} \frac{N_w}{T_w} T_n T_k T_h T_w$$
$$DV_{...,ht,kt}^{Ker} = \frac{N_n}{T_n} \frac{N_k}{T_k} \frac{N_c}{T_c} \frac{N_r}{T_r} \frac{N_s}{T_s} \frac{N_w}{T_w} \frac{N_h}{T_h} T_k T_c T_r T_s$$
$$DV_{...,ht,kt}^{In} = \frac{N_n}{T_n} \frac{N_c}{T_c} \frac{N_r}{T_r} \frac{N_s}{T_s} \frac{N_w}{T_w} \times$$
$$T_n T_c (N_h + T_r - 1)(T_w + T_s - 1)$$
$$DV_{...,ht,kt} = DV_{...,ht,kt}^{Out} + DV_{...,ht,kt}^{Ker} + DV_{...,ht,kt}^{In}$$

For permutation $\langle \{nt, ct, ht, wt, rt\}, st, kt \rangle$

$$DV_{...,st,kt}^{Out} = 2 \frac{N_n}{T_n} \frac{N_k}{T_k} \frac{N_c}{T_c} \frac{N_r}{T_r} \frac{N_s}{T_s} \frac{N_h}{T_h} \frac{N_w}{T_w} T_n T_k T_h T_w$$
$$DV_{...,st,kt}^{Ker} = \frac{N_n}{T_n} \frac{N_k}{T_k} \frac{N_c}{T_c} \frac{N_r}{T_r} \frac{N_s}{T_s} \frac{N_w}{T_w} \frac{N_h}{T_h} T_k T_c T_r T_s$$
$$DV_{...,st,kt}^{In} = \frac{N_n}{T_n} \frac{N_c}{T_c} \frac{N_r}{T_r} \frac{N_h}{T_h} \frac{N_w}{T_w} \times$$
$$T_n T_c (T_h + T_r - 1)(T_w + N_s - 1)$$
$$DV_{...,st,kt} = DV_{...,st,kt}^{Out} + DV_{...,st,kt}^{Ker} + DV_{...,st,kt}^{In}$$

For permutation $\langle \{nt, ct, ht, wt, st\}, rt, kt \rangle$

$$DV_{...,rt,kt}^{Out} = 2 \frac{N_n}{T_n} \frac{N_k}{T_k} \frac{N_c}{T_c} \frac{N_r}{T_r} \frac{N_s}{T_s} \frac{N_h}{T_h} \frac{N_w}{T_w} T_n T_k T_h T_w$$
$$DV_{...,rt,kt}^{Ker} = \frac{N_n}{T_n} \frac{N_k}{T_k} \frac{N_c}{T_c} \frac{N_r}{T_r} \frac{N_s}{T_s} \frac{N_w}{T_w} \frac{N_h}{T_h} T_k T_c T_r T_s$$
$$DV_{...,rt,kt}^{In} = \frac{N_n}{T_n} \frac{N_c}{T_c} \frac{N_s}{T_s} \frac{N_h}{T_h} \frac{N_w}{T_w} \times$$
$$T_n T_c (T_h + N_r - 1)(T_w + T_s - 1)$$
$$DV_{...,rt,kt} = DV_{...,rt,kt}^{Out} + DV_{...,rt,kt}^{Ker} + DV_{...,rt,kt}^{In}$$

**Innermost** $nt$ **and** $ct$: As discussed above, choosing $nt$ or $ct$ as the second loop in $\langle \dots, kt \rangle$ is inferior to choosing one of $\{wt, ht, st, rt\}$. A similar argument can be used to establish that choosing $nt$ or $ct$ as the innermost loop is inferior to choosing one of $\{wt, ht, st, rt\}$. The only difference between the two arguments is that now all cost functions have an extra factor $\frac{N_k}{T_k}$ (since $kt$ is not the innermost loop anymore), but the rest of the reasoning still applies. Thus, no additional cases arise to be solved.

**Summary:** By analyzing the algebraic structure of the cost expressions, as described above, we have identified that only eight equivalence classes of tiling permutations need to be considered:

$$\langle \{kt, ct, rt, st\}, \{nt, ht\}, wt \rangle \qquad \langle \{kt, ct, rt, st\}, \{nt, wt\}, ht \rangle$$
$$\langle \{nt, kt, ht, wt\}, \{ct, rt\}, st \rangle \qquad \langle \{nt, kt, ht, wt\}, \{ct, st\}, rt \rangle$$
$$\langle \{nt, ct, ht, rt, st\}, wt, kt \rangle \qquad \langle \{nt, ct, wt, rt, st\}, ht, kt \rangle$$
$$\langle \{nt, ct, ht, wt, rt\}, st, kt \rangle \qquad \langle \{nt, ct, ht, wt, st\}, rt, kt \rangle$$

Only one arbitrary representative permutation from each set is selected for further analysis, since all elements in the set have exactly the same cost expression for data movement. Thus, the search space is drastically reduced from 5040 distinct tile-loop permutations to only 8 cases for single-level tiling, and $8^L$ cases for $L$-level tiling instead of $5040^L$ cases.

# 5 MULTI-LEVEL TILE-SIZE OPTIMIZATION

In this section, we present our approach to optimizing multi-level tiled CNN. Due to the multiple levels of cache on multiprocessors, multi-level tiling is beneficial to optimize data movement at the different levels in the memory hierarchy. In general, while cache capacities at later levels increase, the bandwidth for data movement between adjacent levels in the hierarchy decrease. Thus the overhead (in time) to move data between different levels in the memory hierarchy will be different. Assuming that concurrent data transfers (of different data) can occur between different levels of the memory hierarchy, we seek to minimize the maximum bandwidth-scaled data-volume across all levels.

For $L$-level tiling, the number of tile parameters will be $7L$, seven tile sizes per level. Since the tiled execution corresponds to a $7L$ loop nest, the range of execution for any iterator $j$ at tile-level $l$ will be $T_j^{l+1}$, i.e., the tile-size for that loop variable at the next outer tiling level, and $N_j$ for the outer-most tile. In the previous section, the data volume expressions for single-level tiling featured ratios of the problem size over the tile size along the different iteration space dimensions, $N_j/T_j$. For multi-level tiling, the expressions will have terms of the form $T_j^{l+1}/T_j^l$, i.e., the expressions for each level involve parametric tile sizes for that tile level and the next outer tile level.

Let $BW^l$ represent the bandwidth available for data transfers and $DV^l$ the volume of data moved between levels $l$ and $l+1$ in the memory hierarchy. We seek a tile configuration that minimizes $max_l \frac{DV^l}{BW^l}$. However, although several publicly available nonlinear solvers can be used to solve the optimization problem developed in the previous section for single-level tiling, none can directly solve a constrained $min(max())$ nonlinear optimization problem. Hence we use the following approach to solve the $L$-level tile optimization problem: solve $L$ constrained optimization problems, where the parametric data volume expression for each level $l$ is minimized in one of those. For the instance of the minimization problem for level $l$, additional constraints are added to the effect that $\frac{DV^l}{BW^l}$ must be greater than or equal to $\frac{DV^k}{BW^k}$ for $k \neq l$.

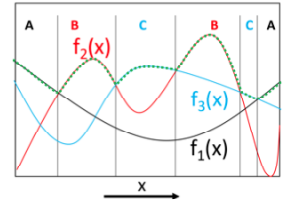

**Figure 3: Example to illustrate approach to multi-level tile size optimization**

Our approach to multi-level tile optimization is illustrated by a simpler example of one-dimensional functions. Fig. 3 shows three functions: $f_1(x)$ (colored black), $f_2(x)$ (colored red), and $f_3(x)$ (colored blue). Consider the problem of finding $min(max(f_1, f_2, f_3))$, where analytical expressions as a function of variable $x$ are available for $f_1$, $f_2$, and $f_3$. We need to find the minimum of the function $f_{comp}$, shown by the dotted line in Fig. 3, but no analytical expression is available for $f_{comp}$ that can be input to a constrained non-linear optimization solver. We solve the $min$-$max$ problem by solving three separate $min(f)$ problems, over the three regions $A$, $B$, and $C$, respectively. $A$ is the region over $x$ where function $f_1$ is greater than or equal to $f_2$ and $f_3$. Similarly, $B$ and $C$ represent regions over $x$ where $f_2$ and $f_3$, respectively, are greater than or equal to the other two functions. The minimum value of $f_{comp}$ over the full range of $x$ can be expressed as $min(m_1, m_2, m_3)$, where $m_1 = min_A(f_1(x))$, $m_2 = min_B(f_2(x))$, $m_3 = min_C(f_3(x))$.

In order to solve for

$$m_{123} = min(max(f_1(x), f_2(x), f_3(x))), X_{lo} < x < X_{hi}$$

we can solve three minimization problems, one each for regions over which the corresponding function has the highest value (regions respectively marked $A$, $B$, and $C$ in Fig. 3):

$$m_1 = min(f_1(x)), f_1(x) \geq f_2(x), f_1(x) \geq f_3(x), X_{lo} < x < X_{hi}$$
$$m_2 = min(f_2(x)), f_2(x) \geq f_1(x), f_2(x) \geq f_3(x), X_{lo} < x < X_{hi}$$
$$m_3 = min(f_3(x)), f_3(x) \geq f_1(x), f_3(x) \geq f_2(x), X_{lo} < x < X_{hi}$$

and then selecting $m_{123} = min(m_1, m_2, m_3)$.

# 6 MICROKERNEL DESIGN FOR CNN

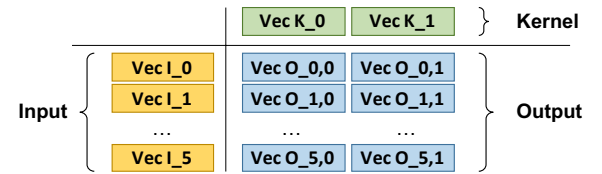

**Figure 4: Conceptual view of outer product scheme**

Along with data movement optimizations, optimizing the throughput of compute-units is critical for achieving close to peak performance. The principal computations in convolutions can be realized using the Fused-Multiply-Add (FMA) operator, which can

```
for(c in 0 to Tc)
 for(r in 0 to Tr)
  for(s in 0 to Ts)
   // outer product
   for(hw in 0 to 6)
    for (k in 0 to 16)
     // FMA
```

**Listing 4: Loop structure of the micro kernel**

```
// Before parallelization
for ( i4 = 0; i4 < Ni; i4 += Ti3)
 for ( j4 = 0; j4 < Nj; j4 += Tj3)
  for ( i3 = i4; i3 < i4 + Ti3 ; i3 += Ti2)
   for ( j3 = j4; j3 < j4 + Tj3 ; j3 += Tj2)
// After parallelization
for ( i4 = 0; i4 < Ni; i4 += Ti3)
 for ( j4 = 0; j4 < Nj; j4 += Tj3)
  for( ip = i4 + tid /(Tj3/PTj3) * Tip; ip< i4+Ti3;
   ip+=(Ti3/PTi3)* Tip)  // parallel
   for( jp = j4 + tid %(Tj3/PTj3) * Tjp; jp< j4+Tj3;
    jp+=(Tj3/PTj3)* Tjp)  // parallel
    for (i3 = ip; i3 < ip + Tip ; i3 += Ti2)
     for ( j3 = jp; j3 < jp + Tjp ; j3 += Tj2)
```

**Listing 5: Loop structure before and after parallelization**

be efficiently executed by the SIMD (vector) units in modern processors. Each core in our benchmarking machines contains two AVX2 (256 bits == 8 floats) SIMD units, which can achieve a combined throughput of $2 \times 8$ FMA operations (16 FMA ops), and has a latency of 4 to 6 clock cycles. The amount of parallelism required to fully utilize the SIMD pipeline can be computed using Little's Law as $latency \times throughput = 6 \times 16 = 96$. Note that these operations should not carry any dependencies.

An outer product scheme, similar to BLIS[24], is used to achieve the required parallelism. Figure 4 shows the conceptual view of our outer product scheme. The output feature is distributed across the vector lanes. In AVX2, each vector register can hold eight single-precision floating-point elements. Two such registers are used to hold the *kernel* elements. Six vector registers, each of which holds a single input image point, are populated using vector broadcasts. The outer product of these six vector registers and two kernel registers are computed using efficient vectorized Fused Multiply Add (FMA) instructions and stored in twelve vector registers. Listing 4 shows the loop structure of our micro-kernel. The actual implementation of the entire microkernel, including loops, is implemented using x86 assembly code.

**Packing:** Efficient vectorization requires stride-1 access along the vectorization dimension. Our scheme vectorizes the output feature dimension ($T$). However, since the kernel layout is $[K, C, R, S]$, $K$ is not the fastest varying dimension. Hence a data layout transformation is performed to make $K$ the fastest varying dimension before the convolutions are processed. We split the dimension $K$ into vector-length sized chunks, and each chunk is laid out contiguously in memory ($[K, C, R, S] \rightarrow [K/VecLen, C, R, S, VecLen]$). Our code generator automatically generates the packing code and this packing cost is included in all experiments.

## 7 OPTIMIZING FOR PARALLELISM

We describe how the sequential cost model is adapted to handle tiled parallel execution. We assume that each core owns a set of private caches (typically L1 and L2) and collectively shares a set of shared caches (typically L3). Since the L3 cache is shared, parallelizing loops that iterate over L3 tiles will cause cache interference. Loops that iterate over L2 tiles as well as loops that iterate over L1 tiles can be parallelized without cache interference. But parallelizing L1 loops will reduce data locality within L2 tiles. Further, parallelizing L2 tile loops achieve coarser parallelism, with lower scheduling overheads. Hence we sub-tile L2 tiling loops to create two-loop bands. Listing 5 shows the tile structure before and after parallelization of a 2D loopnest. The outermost band ($ip$ and $jp$) is used for parallelization and the inner band ($i3$ and $j3$) is executed sequentially by each core. Parallelizing certain dimensions like $W$ and $H$ will result in write conflicts. While these conflicts can be avoided by using atomic operations or synchronizations, the overhead is high. Hence, our model only considers parallelism along the non-reduction dimensions. The cost modeling in the parallel case is very similar to the sequential cost model explained in Sec. 5; hence we only describe the differences in this section. Even though the memory-to-L3 data movement remains the same, the effective bandwidth may be higher in the parallel case. Hence, we use a synthetic benchmark to determine the parallel memory-to-L3 bandwidth and use this bandwidth in the cost model. The parallel L3-to-L2 data movement cost may also change as the available L3 bandwidth is split across multiple cores. The per-core L3-to-L2 bandwidth is also computed using synthetic benchmarks. The parallel L3-to-L2 cost computation is similar to the cost computation explained in Sec. 5 and can be obtained by replacing $T\alpha 3$ in with $PT_\alpha 3$ where $\alpha \in non\ reduction\ dimensions$. $T\alpha 3/PT\alpha 3$ is the amount of parallelism along dimension $\alpha$. A constraint is added to ensure that the total amount of parallelism is equal to the total number of cores ($\prod T\alpha 3/PT\alpha 3 == num\_cores$). The rest of the constraints remain the same. The L2-to-L1 bandwidth and L1-to-register bandwidth used in the parallel case is the same as the sequential case. The parallel cost model is then solved using the same min-max formulation from Sec. 5.

## 8 PUTTING IT ALL TOGETHER

In this section, we discuss some aspects of the overall process for generation of optimized CNN code that have not been previously described. We first demonstrate the way to handle parallel execution and then present the work flow of the full optimization system.

**System Workflow:** The design of the microkernel (Section 6) is entirely dictated by the latency and throughput of the FMA units and is not dependent on the cache or memory parameters. Hence, for a given machine, the same micro-kernel is used for all problem sizes. However, the tile sizes and loop permutation of the loops surrounding the microkernel is dependent on the problem specification. Algorithm 1 shows an overview of our permutation and tile-size selection process. Function *GetPrunedPermutation* returns the set of pruned permutations. The loop at line 3 iterates over each permutation and finds the best tile-sizes for the given permutation. For a given permutation ($pm$), we initialize the *FixedTileSizes* as an empty array at line 5, we first find the tile-sizes for the most-constrained level and fix the tile size corresponding to this level.

**Input** : ProblemSize, HardwareSpec
**Output**: LoopPermutation, TileSize

1 PrunedPermuSet ← GetPrunedPermutations ();
2 GlobalSoln.Cost ← INT_MAX;
3 **for** pm ∈ PrunedPermuSet **do**
4     NotVisitedLvls ← [Reg, L1, L2, L3];
5     FixedTileSizes ← [] ;
6     **while** NotVisitedLvls ≠ ∅ **do**
7         MinCost ← INT_MAX;
8         **for** ObjLvl ∈ NotVisitedLvls **do**
9             [CurCost, CurTileSizes] ←
            ArgMinSolve (ProblemSize, HardwareSpec, ObjLvl,
            pm, FixedTileSizes, NotVisitedLvls);
10             **if** MinCost > CurCost **then**
11                 MinTileSizes ← CurTileSizes;
12                 MinLevel ← ObjLvl;
13                 MinCost ← CurCost;
14             **end**
15         **end**
16         NotVisitedLvls.remove (MinLevel) ;
17         FixedTileSize.add ( getTileSizeforLevel (MinTileSizes, MinLevel) ) ;
18     **end**
19     **if** MinCost < GlobalSoln.Cost **then**
20         GlobalSoln ← {pm, FixedTileSize, MinCost}
21     **end**
22 **end**
23 IntegerSoln ← FloorToInteger (GlobalSoln);
24 FinalSolution.TileSizes ← LoadBalancer (finalIntegerSol);
25 **return** [FinalSolution.pm, FinalSolution.TileSizes]

**Algorithm 1:** Permutation and Tile Selection Algorithm

Next, among the remaining levels, we find the tile-sizes for the most-constrained level and find the tile-sizes for that level. This process is repeated until the tile-sizes for all levels are computed. However, the cost of each level is not known a priori. The maximum constraining level is found using the following steps. For each level: (i) add a constraint to mark the current level as the most constraining one, (ii) invoke the solver to find the tile-sizes which minimizes the cost under the former constraint, (iii) select the level with the minimum cost (min-max formulation). Each iteration of loop at line 6 represents this computation. The loop at line 8 finds the minimum cost assuming that the current level (*ObjLvl*) is the level with maximum constraints. Line 9 invokes the Ipopt solver[37] by setting the constraint that the *ObjLvl* is the most constrained level. The *if* condition at line 10 keeps track of the minimum cost and the associated level. The tile sizes for the most constrained level are then fixed and removed from the search space (lines 16–17). Function *getTileSizeforLevel* is a helper function to extract the tile-sizes for a given level. This entire process is repeated for each permutation to find the best permutation and tile-sizes. Note that the tile-sizes returned from the solver are real numbers; however, tile-sizes should be integers. We *floor* each tile-size to obtain the integer solution. The tile sizes are then adjusted to minimize the core idling (load balance).

## 9 MODEL VALIDATION

We present our experimental evaluation in two parts: first, in this section we discuss model validation, followed in the next section by a comparison with state-of-the-art alternatives: Intel oneDNN [25] and AutoTVM [6, 40].

For our experimental evaluation, we used all CNN benchmarks used by TVM in the extensive comparative evaluation [6] against various other CNN optimization frameworks. The benchmarks used by TVM include all twelve conv2d operators from Resnet-18[13], and the nine depth-wise conv2d operators from MobileNet[14]. In addition we used all eleven conv2d operators from Yolo-9000[29]. All benchmark parameters are shown in Table 1. All input and output tensors were stored in NCHW layout and all kernel tensors were stored in KCRS layout. Any time expended in internal layout transformations was included in the measured execution time for all codes.

The experiments described in this section were performed by measuring single-core performance and profiling hardware counters on an 8-core Intel Core i7-9700K CoffeeLake processor, with 32KB L1 cache per core, 256KB L2 cache per core, and a shared 12MB L3 cache. Hardware counter events were profiled by use of Likwid [33].

For each of the 32 conv2d operators, a sampling of the space of tile-size combinations was performed to select around 100 configurations uniformly distributed in the full space of tile-size combinations. For each code configuration, we generated the model-predicted score, measured performance by executing it, and gathered hardware counter events for data movement volume at the register, L1 cache, L2 cache, and L3 cache levels.

We sought to answer the following questions:

(1) Given a set of alternative tile configurations for a benchmark, how does the rank ordering of those code configurations by use of the analytical model compare with that based on measured performance? The rationale for such an assessment is that the effectiveness of a compiler performance model in differentiating between configurations is much more important than the absolute error between modeled execution time and measured execution time.

(2) How does the rank ordering of code configurations by the model compare with the measured data volumes at the different levels of the memory hierarchy?

(3) What is the loss-of-performance for a model-selected configuration when compared to the best performing configuration in the sampled set? We evaluated a top-1, top-2 and top-5 loss-of-performance score, where top-k means the best performance among the top k predicted configurations by the model.

Figure 5 presents the loss-of-performance comparing model-predicted best configurations and the actual best among the 100 or so configurations evaluated for each benchmark. For each conv2d operator, we calculated three loss ratios. The top-one one represents the loss of performance of the best-predicted case by our model over the actual best code version. The top-two loss represents the loss of performance of the better of the top-2 versions predicted by the model over the actual best code version. For the top-five loss, we take the best among the top 5 cases based on prediction. Our experiment shows that for all thirty-two conv2d operators, the model predicted best code versions always achieve less than 4.5% loss , i.e., the model always finds a code version that achieves 95.5% performance compared to the actual best code version in the sampled configuration space. For most operators (thirty of thirty-two), the loss is less than 3%.

**Table 1: Configurations of conv2d operators in Yolo-9000 (left), ResNet-18 (middle) and MobileNet (right); K: # output channels; H, W: input image height and width; C: #input channels; R/S kernel size; batch size = 1; kernel stride = 1/2 (2 if marked with * after kernel name, 1 otherwise)**

| Layer | K | C | H/W | R/S |
|-------|-----|-----|------|-----|
| Y0 | 32 | 3 | 544 | 3 |
| Y2 | 64 | 32 | 272 | 3 |
| Y4 | 128 | 64 | 136 | 3 |
| Y5 | 64 | 128 | 136 | 1 |
| Y8 | 256 | 128 | 68 | 3 |
| Y9 | 128 | 256 | 68 | 1 |
| Y12 | 512 | 256 | 34 | 3 |
| Y13 | 256 | 512 | 34 | 1 |
| Y18 | 1024 | 512 | 17 | 3 |
| Y19 | 512 | 1024 | 17 | 1 |
| Y23 | 28269 | 1024 | 17 | 1 |

| Layer | K | C | H/W | R/S |
|-------|-----|-----|------|-----|
| R1* | 64 | 3 | 224 | 7 |
| R2 | 64 | 64 | 56 | 3 |
| R3 | 64 | 64 | 56 | 1 |
| R4* | 128 | 64 | 56 | 3 |
| R5* | 128 | 64 | 56 | 1 |
| R6 | 128 | 128 | 28 | 3 |
| R7* | 256 | 128 | 28 | 3 |
| R8 | 256 | 128 | 28 | 3 |
| R9 | 256 | 256 | 14 | 3 |
| R10* | 512 | 256 | 14 | 3 |
| R11* | 512 | 256 | 14 | 1 |
| R12 | 512 | 512 | 7 | 3 |

| Layer | K | C | H/W | R/S |
|-------|-----|-----|------|-----|
| M1 | 32 | 32 | 112 | 3 |
| M2* | 64 | 64 | 112 | 3 |
| M3 | 128 | 128 | 56 | 3 |
| M4* | 128 | 128 | 56 | 3 |
| M5 | 256 | 256 | 28 | 3 |
| M6* | 256 | 256 | 28 | 3 |
| M7 | 512 | 512 | 14 | 3 |
| M8* | 512 | 512 | 14 | 3 |
| M9 | 1024 | 1024 | 7 | 3 |

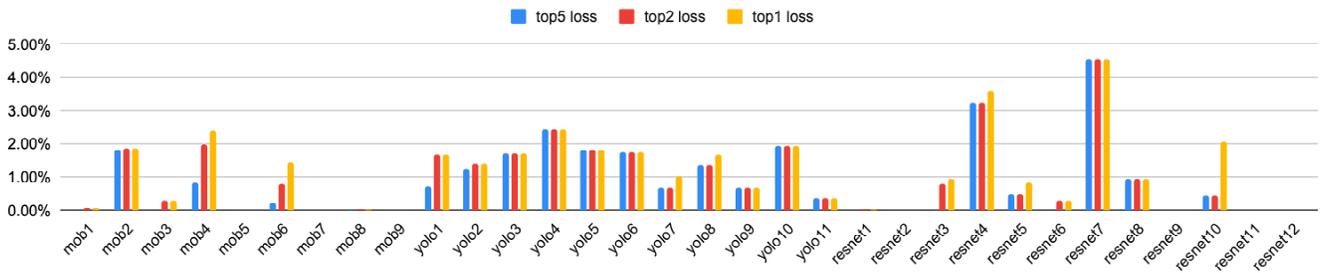

**Figure 5: Model prediction performance loss over 100 grid sampling for Mobilenet, Yolo-9000, and Resnet-18 on i7-9700K**

Figure 6 shows the correlation of predicted performance with actual performance and data movement hardware counters (registers, L1, L2, and L3) for three of the benchmarks:Resnet-9, Mobnet-2, and Yolo-5. Each of the three columns of graphs in the figure correspond to one of those three conv2d operators. In these graphs, the Y-axis represents one of the following metrics: Performance (GFLOPs), number of register load/stores, and L1/L2/L3 cache misses, one chart for each metric, in that order from top to bottom. The different configurations are ordered from left to right along the X-axis on the basis of model-predicted performance, with the best-predicted case at the left end, and the worst-predicted case at the right end.

The first row of charts shows that there is a strong correlation between actual performance and predicted performance.- code versions with higher performance generally also have higher model-predicted scores. The other plots shows a strong correlation between data movement hardware counter measurement for the predicted bottleneck resource and the predicted performance. Since the predicted performance is based on the predicted bottleneck resource, we would expect correlation with hardware counter measurements for that resource. For both Resnet9 (left column) and Mobnet2 (middle column), the model predicts that the register level is the most constraining one. Indeed, the experimental measurements show a strong correlation with hardware counter measurements of load/stores. It is interesting to note that for both benchmarks there is no correlation with hardware counter measurements at some other levels, specifically L1 and L3. Both registers and L3 are predicted to be constraining resources for Yolo5 (right column) and this is also seen in the experimental data.

## 10 COMPARISON WITH STATE-OF-THE-ART LIBRARY AND AUTO-TUNING

In this section, we present a comparative experimental evaluation of the code generated by MOpt with a state-of-the-art library (Intel oneDNN [25]) and a state-of-the-art auto-tuning system (AutoTVM [6, 40]). The experiments were carried out on two systems: an 8-core Intel Core i7-9700K CoffeeLake processor, with 32KB L1 cache per core, 256KB L2 cache per core, and a shared 12MB L3 cache and an 18-core Intel i9-10980XE CascadeLake processor, with 32KB L1 cache per core, 1MB L2 cache per core, and a shared 24.75MB L3 cache.

We compare the performance of code generated by MOpt with two state-of-the-art frameworks: (i) Intel oneDNN (v1.5) library, and (ii) TVM (v0.6). TVM relies on auto-tuning and machine learning models to generate efficient code. All MOpt codes and oneDNN were compiled using the Intel ICC 2019 compiler with flags "-O3 -march=native -qopenmp". TVM recommends using the LLVM framework; hence we used LLVM-8. TVM tuning was based on their recommended template: "generic.schedule_conv2d_nchw"[38]. We used XGBTuner as the ML tuning model, and we set "LLVM -mcpu=core-avx2 or -mcpu=skylake-avx512" based on the target to ensure that the generated code was vectorized for the appropriate ISA (avx2 for i7, avx512 for i9). For each CNN benchmark, we ran TVM's auto-tuner with its internal ML model to find the best configuration over 1000 trials.

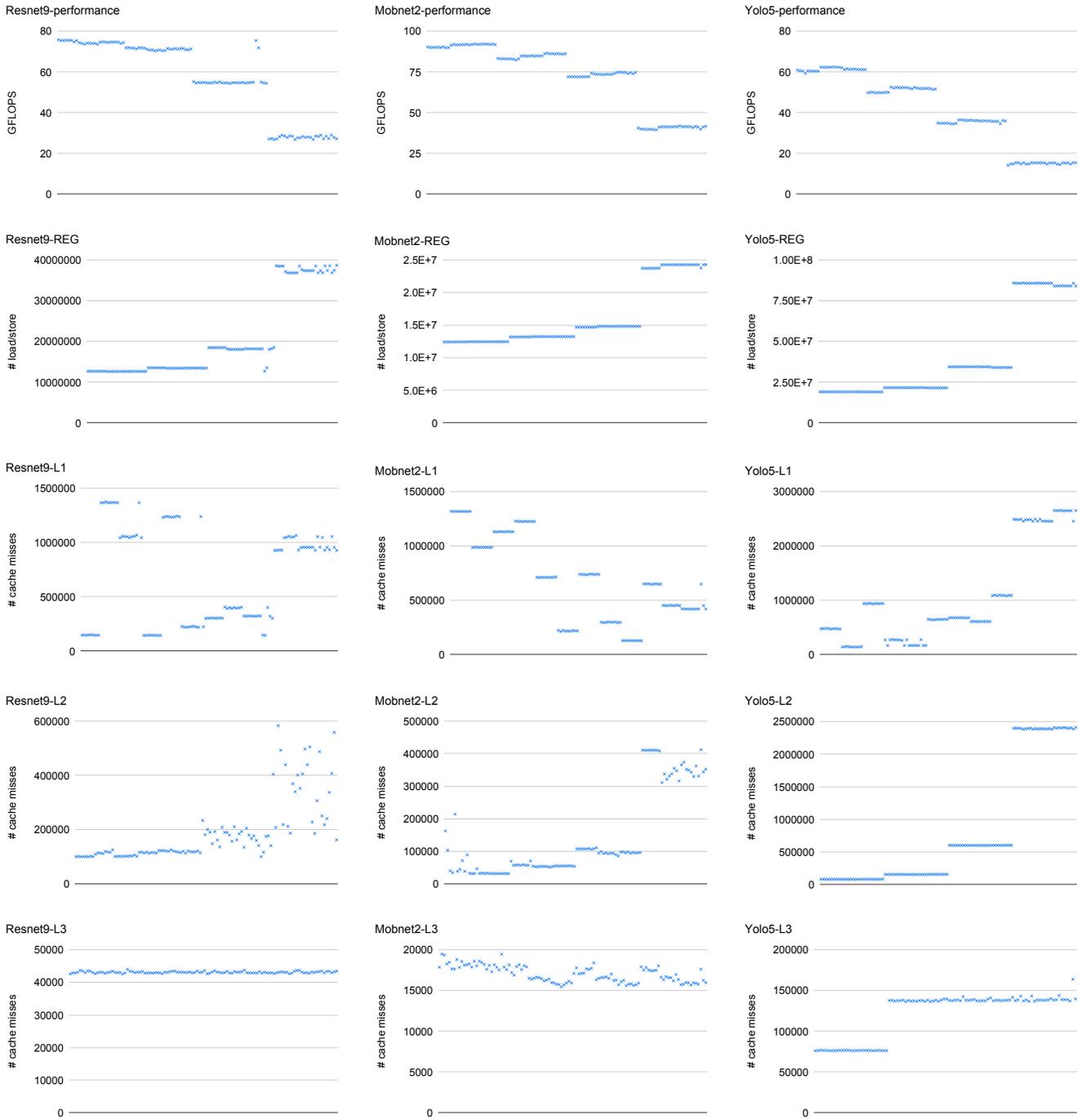

**Figure 6: Model-predicted rank ordering versus actual measurement on i7-9700K. Left: Resnet9, Middle: Mobnet2, Right: Yolo5; Top: Performance (GFLOPS), followed by Reg. load/stores, L1 misses, L2 misses, L3 misses. Points are ordered along X-axis in decreasing order of predicted performance.**

We compare TVM and oneDNN agaist two MOpt code versions (i) MOpt-1: A single code version generated with the configuration with minimum modeled cost and (ii) MOpt-5: Five code versions were synthesized based on the top 5 modeled configurations. The reason we also include MOpt-5 is to highlight the potential for performance improvement by inclusion of limited empirical auto-tuning to MOpt. Since the modeling in MOpt is based on an idealized fully associative cache, occasionally we find (e.g., Yolo9 and Yolo18) that conflict misses cause a significant drop in performance. But when we consider the top five configurations generated by the MOpt framework, it turns out that these configurations rarely experience pathological conflict miss scenarios and the best among the top five performs very well.

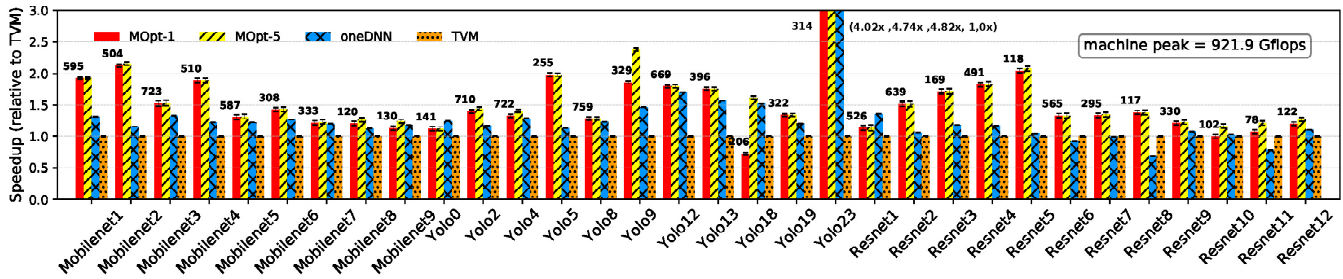

**Figure 7: Performance (relative to TVM) and variance for Mobilenet, Yolo-9000, and Resnet-18 on i7-9700K**

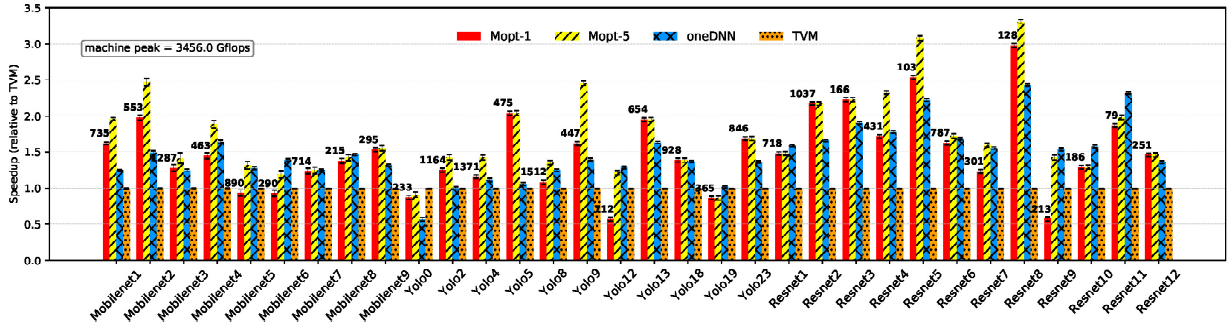

**Figure 8: Performance (relative to TVM) and variance for Mobilenet, Yolo-9000, and Resnet-18 on i9-10980XE**

We repeated each experiment 50 times on the system, using 8 threads on i7-9700k and 16 threads on i9-10980xe. We excluded the very first run since it often includes additional time for loading libraries. In order to avoid cache reuse across successive runs, we flushed the cache between runs and measured the execution time of each run individually. We turned off DVFS and turbo-boost, and locked the clock at base frequency to reduce the variability across runs. For each benchmark, we report mean GFLOPS achieved over 50 runs. The bar charts and the left vertical axes in Figure 7 show the performance, normalized to TVM's performance. As recommended by a popular approach for statistically-rigorous performance measurements [10], we also report the 95% confidence interval. The interval is shown on top of each bar, as a characterization of variance; in some cases, it is so small that it is barely visible. We also show the actual GFLOPS value of the MOpt-based code above the corresponding bar.

The geometric means of speed-up of MOpt over oneDNN are: On i7-9700k, 1.16x on the Yolo, 1.37x on the ResNet, and 1.24x on MobileNet. On i9-10980xe, 1.26x on the Yolo, 1.08x on the ResNet, and 1.14x on MobileNet. The geometric means of speed-up of MOpt over TVM are: On i7-9700k,1.73x on the Yolo, 1.40x on the ResNet, and 1.52x on MobileNet. On i9-10980XE, 1.53x on the Yolo, 1.84x on the ResNet, and 1.56x on MobileNet.

## 11 RELATED WORK

**Tile size optimization:** Some previous research has focused on tile size optimization based on analytical modeling [32, 39]. However, they relied on heuristic search. Recently, Li et. al [21] developed an analytical modeling approach and its solution using nonlinear solvers, for optimizing data movement for tensor contractions. However, their work only addressed sequential computing and was restricted to tensor contractions and could not be applied to CNNs. Renganarayana et. al[30] developed a framework based on integer geometric programming to optimized tile size selection if the optimization problem could be expressed as a polynomial. While our one-level tile-size optimization formulation is a posynomial, the constraints arising in the multi-level tile optimization problem are no longer polynomials.

Some other previous efforts have formalized the tile size selection problem as a constrained optimization problem. Sarkar et. al [31] presented a model for optimizing memory cost for doubly nested loops, and limited the dimension of loop nest to not greater than three. Krishna [17] et. al utilized a nonlinear solver to find optimal tile sizes to minimize disk I/O for tensor contraction, but they only addressed on single level of tiling. Cociorva et. al [7] proposed a model for optimizing inter-processor communication under memory constraints, restricted to tensor contraction. Lin et. al [22] developed a tool that used a convex solver to optimize tile size for direct buffer access. However, it relied on heuristic search to find loop permutations and did not comprehensively cover the full loop permutation space, and they also only addressed a single level of tiling.

**Polyhedral compilers:** Polyhedral compilers such as Polly[11], Pluto[5], PPCG [36] perform tile sizes optimization and loop parallelization based on the polyhedral model. Tensor Comprehension[35] is an automatic compiler for converting tensor computations to high-performance machine learning kernels based on the polyhedral model. However, a fundamental limitation of polyhedral compilers is that the cost models used for optimization are linear. The tile-size optimization problem is inherently non-linear. Polyhedral compilers are forced to separate tile-size optimization from tile-loop permutation and therefore have not demonstrated code generation for CNN whose performance matches vendor library code (like Intel oneDNN) or optimizers that use auto-tuning (like TVM).

**Specialized Machine Learning compilers:** PlaidML [27] is a portable tensor compiler that compiles deep learning codes on mobile devices. It automatically applies tiling transformation to improve efficiency of training. XLA (Accelerated Linear Algebra) [19] is a domain-specific compiler that improves performance for linear Algebra operators inside Tensorflow[1]. XLA fuses Tensorflow operators in the same graph, so it reduces the requirements to write intermediate values and number of kernel calls. TVM [6] is an automatic end-to-end optimizing compiler for improving the performance of deep learning systems. It works with deep learning frameworks like Pytorch[26] and Keras[15] and supports code generation for different hardware platforms. It extends and uses Halide [28] as its internal representation. Its optimization is driven by an ML-based cost model that trains itself by using auto-tuning data collected when running on the target platform. It has been demonstrated to achieve much higher performance than other existing CNN optimizing frameworks like PPCG, PlaidML, XLA, etc. [6, 40]. Thus, TVM represents the current state-of-the-art in CNN optimization. In this paper, we therefore compare performance with it.

**CNN libraries:** Intel's oneDNN[25] is a state-of-the-art optimized neural network library for Intel Architectures. We have compared performance with oneDNN.

## 12   DISCUSSION

To the best of our knowledge, this paper presents the first demonstration that a purely analytical modeling approach for optimized code generation for CNN can achieve performance comparable to or better than the current state-of-the-art in both optimized vendor libraries and auto-tuning based optimizers that perform actual execution of candidate code versions on the target platform. Further improvement of performance is possible by via incorporating the strengths of these systems into MOpt, as discussed below.

Table 2 contrasts the strengths and limitations of oneDNN, TVM, and MOpt. oneDNN is a highly optimized vendor library that includes highly optimized microkernels developed and optimized by Intel engineers over many years. However, it dynamically chooses among a small number of pre-determined tiled code structures based on the CNN array sizes provided at invocation, i.e., it performs minimal design-space exploration. TVM performs a search through a limited design space, as specified by the tuning script.

A significant difference between our model-driven search methodology and TVM's auto-tuning based search is the extent of the space that can be effectively explored. Our search time is relatively independent of the problem size, while TVM's search time for a specified number of samples is essentially proportional to the number of operations of the specific CNN modeled. For example, TVM took 1 minute versus 109 minutes to search for the optimal code for the small first stage versus the large last stage of the Yolo-9000 pipeline. However, MOpt only took 9 seconds and 23 seconds, respectively, for optimizing these two problem cases. Therefore a judicious constraining of the full search space is essential for using TVM (as detailed in Sec. 10, we use the script recommended by the developers of TVM), i.e., comprehensive design-space exploration is not practical.

**Table 2: Strengths/limitations of oneDNN, TVM and MOpt**

|         | Auto tuning | Micro Kernel | Design Space Exploration |
|---------|-------------|--------------|--------------------------|
| **oneDNN** | ✗ | Highly optimized | Minimal |
| **TVM** | ✓ | NA | Limited |
| **MOpt** | ✗ | Not highly optimized | Comprehensive |

MOpt's strength is comprehensive design-space exploration to seek tile-loop structures and tile sizes that minimize the data volume at the bottleneck resource in the multi-level cache hierarchy. It does not use any empirical auto-tuning in its search and uses a micro-kernel that is not as highly optimized as oneDNN's. Nevertheless, the achieved performance of MOpt's code on the CNN stages of three DNN pipelines is almost always better and often much better than TVM's code, and comparable and sometimes much better than oneDNN. While data-movement volume is a significant factor that affects performance, other factors are also important, which are very challenging to model, such as conflict misses in real caches with finite set-associativity. A direction for ongoing/future research is to combine our model-driven approach with a limited amount of auto-tuning via actual execution on the target platform. One direction we explored was to incorporate a data-volume-model guided search within TVM's auto-tuning based search. However we faced a fundamental problem: TVM uses LLVM's compiler to generate vectorized code and it performs loop transformations in its backend that we cannot control. The performance of the final resulting code was affected very significantly by the LLVM backend so that a tile loop structure and tile sizes for which MOpt achieves very high performance can produce very low performance through the TVM-LLVM chain because of LLVM's transformations. TVM plans extensions to allow fixed microkernels at the inner-most level instead of the sole current path of LLVM code generation. When that feature is available, we expect to be able to incorporate MOpt's model-driven search into TVM's auto-tuning and gain the combined benefit of comprehensive design-space exploration and empirical auto-tuning.

Further planned work will apply the analytical modeling approach to optimize CNN on other target platforms. GPUs, FPGAs, distributed-memory systems, and accelerator arrays can be abstracted in a similar manner, as hierarchical systems with memory capacity at each level, with consideration for achieving adequate parallelism, leading to multi-level tile-size optimization problems. One important extension will be the modeling of spatial locality. This can be done by adapting the data volume expressions to count the number of cache lines (or DRAM transactions for GPUs): Use $\lceil \frac{T_k}{L} \rceil$ instead of $T_k$, where $L$ is the cache line-size in words and $T_k$ is the tile size along the fastest-varying dimension of an array. This reflects the fact that the movement of data is actually in units of larger granularity—cache lines or fixed-size DRAM transactions (on GPUs)—and not individual elements.

Finally, there is significant potential for application of this model-driven tile-optimization approach to overcome a fundamental limitation of polyhedral compilers: tile size optimization is currently infeasible because parametric tile size variables cause the array indexing expressions to become non-affine and thus out of the scope of the inherent modeling machinery within the polyhedral model. For a significant and practically important subset of matrix/tensor computations, a tile-footprint based cost-modeler and optimizer can be plugged into a polyhedral compiler, enabling iterative search across tile loop permutations and fusions by exposing MOpt-like parametric tile size optimization to guide loop transformations.

## 13 CONCLUSION

We present a new approach to overcome the design-space explosion problem that has thwarted effective compile-time modeling and optimized code generation for CNNs. Although the space of possible configurations is extremely large, we devise an effective analytical modeling approach to search in this space. The structure of data movement cost expressions is exploited to achieve dramatic space pruning. Constrained non-linear optimization problems are used to find multi-level tile sizes that minimize bandwidth-scaled data volume at the most constraining level in the memory hierarchy. Experimental results demonstrate that achieved performance is superior to code generated by TVM and can be comparable to or better than Intel's oneDNN. Further improvements are possible by incorporating better microkernels and by using empirical auto-tuning. The methodology for full design-space exploration and tile-size optimization can also be used to enhance the performance of libraries such as oneDNN, optimizing code generators such as TVM, and polyhedral compilers.

## ACKNOWLEDGMENTS

This work was supported in part by the U.S. National Science Foundation through awards 1946752, 1919122 and 2018016.

## A ARTIFACT APPENDIX

### A.1 Abstract

This artifact describes the steps to reproduce the results presented in this work.

### A.2 Artifact Check-list (Meta-information)

- **Program:** Mopt, TVM, OneDNN
- **Compilation:** Intel C++ compiler, LLVM-10, LLVM-8, Python3.8 (scripts are provided)
- **Benchmark:** conv2d operators in ResNet, MobileNet, and Yolo (bechmarking scripts are provided)
- **Run-time environment:** Linux Ubuntu 18.04 LTS, Miniconda
- **Hardware:** Intel i7-9700k and Intel i9-10980xe CPU

- **Execution:** Execution scripts are provided
- **Metrics:** Execution time/GFLOPS and data movement
- **Output:** Log file with GFLOPS/data movement
- **How much disk space required (approximately)?:** 100GB
- **How much time is needed to prepare workflow (approximately)?:** One hour (based on the dependencies)
- **How much time is needed to complete experiments (approximately)?:** 96 hours
- **Publicly available?:** Yes
- **Code licenses (if publicly available)?:** Custom (provided with artifact)
- **Archived (provide DOI)?:** 10.5281/zenodo.4322031

### A.3 Description

*A.3.1 How to Access.* All the source code, benchmarks, and scripts associated with this work are available under https://doi.org/10.5281/zenodo.4322031. A copy of the software is also maintained at https://github.com/HPCRL/ASPLOS_artifact.

*A.3.2 Hardware Dependencies.* Experiments requires the following CPUs: Intel i7-9700k and Intel i9-10980xe

*A.3.3 Software Dependencies.*

- Python 3.8 (miniconda) wilth amplpy, sympy, joblib modules
- Intel C++ Compiler 2019
- AMPL Ver. 20181102
- IPOPT 3.12
- GCC 7.5
- LLVM version 10.0 (for experiment on avx512 only)
- LLVM version 8.0 (for tvm only)
- Likwid (for hardware counter measurements on i7-9700K)

*A.3.4 Benchmarks.* we use conv2d operators in ResNet-18, MobileNet, and Yolo9000 as the benchmarks.

### A.4 Installation

We recommend installing Miniconda and using a virtual environment for the experiment. Use pip to install the following modules: amplpy, sympy, joblib. Install AMPL binary and IPOPT binary (links below). Install CMake, Intel C++ compiler and LLVM compiler following the official instructions. Mopt's micro-kernel generator can be compiled using cmake (see README.md for additional instructions). Compile TVM v0.6 commit 008aa838139bcd8e66c680f14a944f7af274a33d using LLVM-8 by following the official instructions (see README.md for additional instructions).

Detailed installation instructions can be found in the README.md file. Important links are listed as follows:

- miniconda: https://docs.conda.io/en/latest/ miniconda.html
- AMPL: https://ampl.com/try-ampl/download-a-free-demo/
- IPOPT: https://ampl.com/products/solvers/all-solvers-for-ampl
- Cmake: https://cmake.org/documentation/;
- Intel C++ Compiler: https://software.intel.com/content/www/us/en/develop/tools/oneapi/components/dpc-compiler.html;
- LLVM https://llvm.org/docs/UserGuides.html;

## A.5 Evaluation and Expected Results

We run each conv2d operator 50 times with cache flush for MOpt, OneDNN, and TVM. All the input and output tensors are stored in the 'NCHW' layout, and the kernel tensor is stored in the 'KCRS' layout. Transposing time, if any, is also included in the measured time. We run each benchmark 50 times and report the average GFLOPs. After disabling hyper-threads and fixing the frequency to the processor's base frequency, we expect to see stable GFLOPs among the 50 times runs. The average GFLOPs should be similar to the reported values in the main paper.